\documentclass[sigconf]{acmart}

\usepackage{arydshln}

\copyrightyear{2020} 
\acmYear{2020} 
\setcopyright{acmcopyright}

\acmConference[CIKM '20]{Proceedings of the 29th ACM International Conference on Information and Knowledge Management}{October 19--23, 2020}{Virtual Event, Ireland}

\acmBooktitle{Proceedings of the 29th ACM International Conference on Information and Knowledge Management (CIKM '20), October 19--23, 2020, Virtual Event, Ireland}

\acmPrice{15.00}
\acmDOI{10.1145/3340531.3412722}
\acmISBN{978-1-4503-6859-9/20/10}

\begin{document}

\fancyhead{}

\title{Learning to Profile: User Meta-Profile Network\\
for Few-Shot Learning}

\author{Hao Gong}
\orcid{0000-0002-9879-4595}
\affiliation{
  \institution{Rakuten Institute of Technology, Rakuten, Inc. Tokyo, Japan}
}
\email{hao.gong@rakuten.com}

\author{Qifang Zhao}
\affiliation{%
  \institution{Rakuten Institute of Technology, Rakuten, Inc. Tokyo, Japan}
}
\email{james.zhao@rakuten.com}

\author{Tianyu Li}
\affiliation{%
  \institution{Rakuten Institute of Technology, Rakuten, Inc. Tokyo, Japan}
}
\email{tianyu.li@rakuten.com}

\author{Derek Cho}
\affiliation{%
  \institution{Rakuten Institute of Technology, Rakuten, Inc. Tokyo, Japan}
}
\email{derek.cho@rakuten.com}

\author{DuyKhuong Nguyen}
\affiliation{%
  \institution{Data Science \& AI Department, Rakuten, Inc. Tokyo, Japan}
}
\email{duykhuong.nguyen@rakuten.com}

\begin{abstract}
Meta-learning approaches have shown great success in solving challenging knowledge transfer and fast adaptation problems with few samples in vision and language domains. However, few studies discuss the practice of meta-learning for large-scale industrial applications, e.g., representation learning for e-commerce platform users. Although e-commerce companies have spent many efforts on learning accurate and expressive representations to provide a better user experience, we argue that such efforts cannot be stopped at this step. In addition to learning a strong profile of user behaviors, the challenging question about how to effectively transfer the learned representation and quickly adapt the learning process to the subsequent learning tasks or applications is raised simultaneously. 

This paper introduces the contributions that we made to address these challenges from three aspects. 1) \textbf{Meta-learning model}: In the context of representation learning with e-commerce user behavior data, we propose a meta-learning framework called the Meta-Profile Network, which extends the ideas of matching network and relation network for knowledge transfer and fast adaptation; 2) \textbf{Encoding strategy}: To keep high fidelity of large-scale long-term sequential behavior data, we propose a time-heatmap encoding strategy that allows the model to encode data effectively; 3) \textbf{Deep network architecture}: A multi-modal model combined with multi-task learning architecture is utilized to address the cross-domain knowledge learning and insufficient label problems. Moreover, we argue that an industrial model should not only have good performance in terms of accuracy, but also have better robustness and uncertainty performance under extreme conditions. We evaluate the performance of our model with extensive control experiments in various extreme scenarios, i.e. out-of-distribution detection, data insufficiency and class imbalance scenarios. The Meta-Profile Network shows significant improvement in the model performance when compared to baseline models.
\end{abstract}



\maketitle

\section{Introduction}
Rakuten has one of the largest e-commerce ecosystems in the world, and serves tens of millions of users by providing diverse cross-domain services from online to offline. Like all other e-commerce platforms, a fundamental task is to understand customers and serve better by using the gained knowledge from huge amounts of collected user behavior data through applications, e.g., personalized recommendation, target prospecting, search ranking and customized advertisement systems \citep{hu2013personalized, perlich2014machine, agichtein2006improving}. Traditionally, this is done through intensive feature engineering to create various features of users with machine learning and deep learning algorithms. However, this strategy relies heavily on the quality of data and prior domain knowledge. For the extracted features, as many of them are generated to serve a particular task, this leads to huge challenges when attempting to reuse it for new tasks.

Representation learning has been proposed to expand the scope of applicability of machine learning, and reduce the dependence on feature engineering \citep{bengio2013representation}. In the case of e-commerce, a well learned representation could be task-agnostic and reflect the general interest of a user. And the great success brought by deep learning has made significant advances to support representation learning \citep{radford2015unsupervised}. However, with the increase of input data size and the deepening of the model structure, the cost of model training greatly increases, requiring many iterations and weight updates during the training optimization \citep{sung2018relation}. Meanwhile, to meet the diverse needs of different users and services, we also face an increasing number of unknown tasks. Considering resource limitations, training a large-scale model from scratch for new tasks is impractical for business purposes due to time constraints. Moreover, the standard deep learning does not offer a satisfactory solution to quickly learn new tasks from small sample data, as the achievement of deep learning heavily relies on the fact that it demands large amounts of labeled examples.

The problem of learning how to efficiently and quickly transfer learned knowledge to new tasks with few training examples is referred to as few-shot learning or meta-learning, and has become the subject of a lot of recent studies \citep{Bengio97onthe, Thrun1998}. With human learning, people can usually learn new concepts and skills successfully with only a few examples \citep{lake2015human}. The research literature on meta-learning shows great diversity, spanning from optimization-based learning \citep{Ravi2017OptimizationAA, finn2017model}, memory network-based learning \citep{santoro2016meta, munkhdalai2017meta}, to metric-based learning \citep{vinyals2016matching, snell2017prototypical, sung2018relation}. Among these research, optimization-based and memory network-based learning approaches show many advantages compared to traditional learning methods, but the training examples still need to be slowly learned by the model into its parameters which limited the universality of the model to solving realistic scenario problems \citep{yu2018diverse, sun2019meta}. In contrast, the metric-based approach allows the fast adaptation of new examples through non-parametric models without causing catastrophic forgetting \citep{vinyals2016matching}.

Although previous studies demonstrate the successes of meta-learning with few-shot settings, there are questions that still remain. Can meta-learning be applied for user representation learning for e-commerce applications? For example, can we perform target prospecting with a user's sequential shopping behavioral data? If so, how can we implement the idea to a production environment, and evaluate the model performance appropriately?

In this work, we propose a cross-domain meta-learning framework called Meta-Profile Network to solve the user profiling problem in representation learning with e-commerce user behavior data, which empowers the model quickly adapt to unseen tasks and leverages the experience from model training simultaneously. We extend the idea of the matching network and relation network approaches \citep{vinyals2016matching, sung2018relation} in a realistic scenario. Meanwhile, we propose the time-heatmap encoding strategy and deep network architecture to boost the learning efficiency in user profiling with long-term cross-domain data. 

More importantly, we believe that a good industrial model should not only perform well in terms of accuracy but also have strong robustness and uncertainty performance under extreme conditions. By comparing to baseline models, we conduct three sets of extensive control experiments to test the performance of our model, i.e. uncertainty to out-of-distribution detection, robustness to data insufficiency and class imbalance.

\section{Related Work}
In this section, three directions of research are briefly described from following aspects - encoding strategy, meta-learning and network architecture - that are closely related to our proposed work. 

\subsection{Long-Term User Behavior Representation and Cross-Domain Data at Rakuten}
The development of long-term human behavior is a relatively slow but dynamic procedure, which is subtly affected by the physical, mental, and social activity during the phases of human life \citep{bornstein2014interaction}. For most people, their long-term shopping behavior changes slowly over time. Unlike short-term behavior, which is influenced by many uncertain emergent factors, long-term behavior could be regarded as an internal attribute of humans. Based on this understanding, to better comprehend the user needs, this work assumes that general users have such potential internal behaviors which are changing slowly with the time window moving. Instead of using the short-term behavior data (e.g., 10 days click-through data) to learn the unexpected interests of users, our model uses data from the past one-year time frame to capture users' long-term behaviors. 

To create an effective user representation and alleviate the learning bias from extreme data skew, it is essential to learn from multiple data sources instead of a single data source. At Rakuten, there are over 70 unique businesses including e-commerce, digital content communication and finical tech services, and the cross-use ratio of all services are over 60\% \citep{rakuten_global_website}. For example, Rakuten Ichiba is the leading online shopping website in Japan, Rakuten Travel provides hotel and flight booking services, and Rakuten Points is a point system based on virtual currency which links all of these services. Together these services build up a diverse ecosystem that covers many aspects of users' daily life. Behavior data derived from the ecosystem is very rich and diverse, which allows for many opportunities to improve customer satisfaction and generate business value. To take advantage of this unique ecosystem, this study utilizes data from across these various domains.

However, with the use of long-term and cross-domain data, problems regarding representation learning with computation efficiency on large-scale data have also arisen. Along with the rise in popularity of many embedding techniques and neural networks in recent years, distributed representation \citep{Bengio2013RepresentationLA, lecun2015deep} has gained significant attention and wide applications. Compared to traditional methods that assemble and aggregate user preferences based on prior human knowledge, embedding with carefully picked behavior sequences and sampling methods have shown success in recommendation and similar item discovery \citep{real-time-embeddings}, and improved the data processing efficiency as well. Such encoding-decoding strategies are used to embed sequential user behavior data into low-dimensional representations in this work (details described in Section 3.1).

\subsection{Meta-Learning}
Meta-learning models are trained over a variety of source tasks and optimize the model to adapt for unknown tasks on the learned distribution space of tasks. Few-shot (or one-shot) learning, the alternative term of meta-learning on few-shot (or one-shot) settings, has shown the ability to generate novel behavior based on inference from only one or few training examples which can benefit real-world applications \citep{santoro2016meta, finn2017model, vinyals2016matching, munkhdalai2017meta, snell2017prototypical, sung2018relation}. However, based on our understanding, the model scalability and flexibility limit the effectiveness of the memory network-based methods and optimization-based methods in solving industry problems. The memory network-based methods use specific structures to generalize the stored experience to unseen tasks. Meanwhile, to avoid model overfitting with training data in meta-training phase, optimization-based methods use a shallow neural network which restrict its capability of using deeper structures. To obviate the disadvantages of other meta-learning methods when facing real-world few-shot classification problems, we use a metric-based method in our work.

Our work extends the ideas of the matching network and the relation network. The matching network proposes the idea that training in the same way as testing, which means the training and testing conditions should match \citep{cai2018memory}. It learns and encodes the meta-data to a low-dimension embedding space in a feed-forward way without updating deep network weights, and then maps the pair of sample datasets (a labeled support set and an unlabelled prediction set) to its label by using the embedding. This design can alleviate the requirement of fine-tuning for new task adaptation. The relation network takes this idea one step further. It proposes a two-branch framework to learn the distance metric with embeddings.

\subsection{Multi-Modal Framework and Multi-Task Learning}
Diverse cognition is critical to user understanding. The information from various modalities is characterized by very different statistical properties. In the field of e-commerce, due to the nature of each domain, data from each service can be regarded as unimodal data. Many studies have shown that a multi-modal framework can learn commonalities from multiple modality data and support the task learning, like the semantic representation studies in vision and language \citep{feng-lapata-2010-visual, bruni2014multimodal}. The cross-domain data can be considered as one kind of multi-modal data. Due to the difference in service coverage and distribution of active users in each domain, to tackle such cross-domain data and make sure the model learning ability not be limited by the diversity, the multi-modal framework are used to build the deep neural network in the meta-training phase in our work. The framework shows the tolerance and flexibility to exploit and synthesize the redundant inherent information to reach understanding. The cross-domain data are from three big domains, viz. user demographic data (numerical and categorical data), behavior data of online shopping service (time series sequential data), and point usage data from Rakuten Points service (time series sequential data). All of these data are connected through a common ID system under the Rakuten ecosystem.

Meanwhile, to avoid losing the learning ability of generalization when focused on a particular single task, we introduce multi-task learning to the model to provide an inductive bias, which allows the model not to learn from one task but a distribution space of tasks. As Rich Caruana \citep{caruana1997multitask} described "Multi-task learning improves generalization by leveraging the domain-specific information contained in the training signals of related tasks." Multi-task learning can be used to augment data to address data insufficient problems, such as class imbalance, noisy, sparse, and insufficient samples.

\section{Methodology}
To learn effective user representations to provide better user experiences, we propose a meta-learning framework called Meta-Profile Network, which learns the similarity of embeddings with a two-branch framework. In this section, first we describe the time-heatmap encoding strategy that is used to handle long-term sequential shopping behavior data. Then it is followed by the introduction of Meta-Profile training strategy and the technical details of the deep network.

\begin{figure}[!ht]
    \centering
    \includegraphics[width=\linewidth]{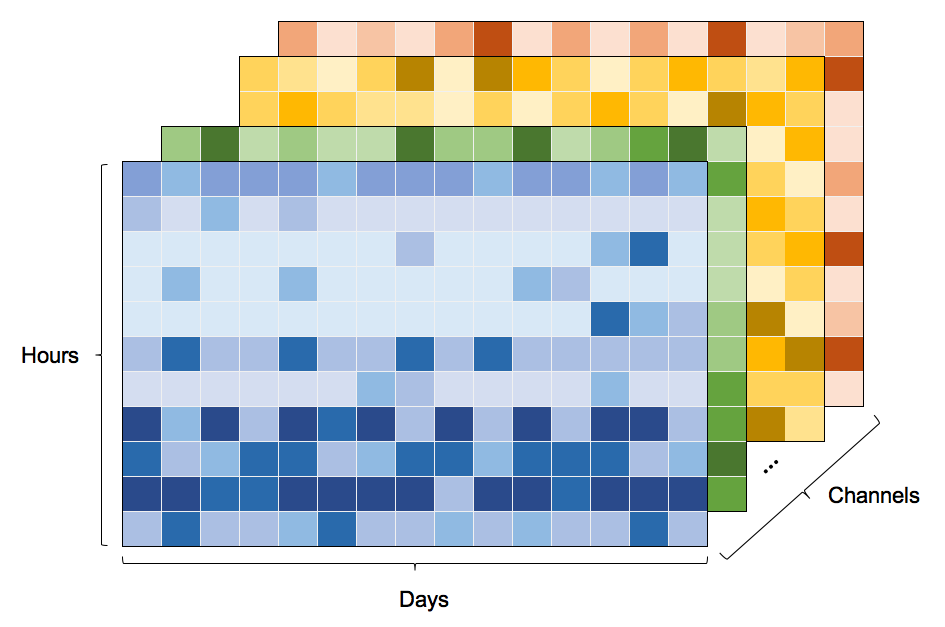}
    \caption{Time-heatmap encoding. Each channel of the 'image' represents the behavior pattern in certain perspective.}
    \label{fig:encoding}
\end{figure}

\subsection{Time-heatmap Encoding Strategy}
At present, the research on image processing with deep learning models has progressed greatly. Inspired by the image data format, we propose the time-heatmap encoding method. As a general digital image is formed by the combination of RGB 3-color channels. Analogous to image data matrices, we use an image-like format to embed the time-series based user behavior data into 'image' channels as shown in Figure.\ref{fig:encoding}. Since we intend to use long-term sequential behavior data (as Table.\ref{tab:raw_data_example} introduced), the time window of data is exactly one year, which can treat equivalently as an image of dimension 24 (hours) × 365 (days). Each channel of the 'image' reveals the behavior pattern in a certain perspective. And the value of each cell in each channel can be any value based on the extraction rule. For example, the first channel represents fashion-related shopping behavior, while the second channel can represent household-related behavior. And the cell value in the first channel can represent the amount that the user spends on purchasing fashion products. We can say each encoded channel is a snapshot of the user's life in the past year from a specific point of view. In general, this encoding strategy gives us the ability to transform the processing of sequential behavior data into image data processing problems. Meanwhile, as a more efficient format, the image-like format reduces the resource requirements for processing large-scale data, standardizes the data size, and enhances the flexibility and scalability of data usage in implementation.

\begin{table}[ht]
\caption{Example of user shopping behavior sequence. The time-heatmap encoding method will convert the raw data from sequence data to image channels.}
\begin{center}
\label{tab:raw_data_example}
\resizebox{\linewidth}{!}{%
    \begin{tabular}{c c c l}
        \toprule
        \textbf{UserID} & \textbf{Price} & \textbf{Datetime} & \textbf{Genre Name by Level} \\
        \midrule
        1 & 2200 & 20XX-XX-XX 21:30:00 & beauty/makeup/lipstick\\
        2 & 3200 & 20XX-XX-XX 21:10:00 & beauty/makeup/lipstick\\
        3 & 2200 & 20XX-XX-XX 14:40:00 & shoes/sandals/straps\\
        4 & 4600 & 20XX-XX-XX 22:40:00 & fashion/outerwear/coat\\
        4 & 1400 & 20XX-XX-XX 20:40:00 & fashion/tops/shirts/blouses\\
        4 & 2000 & 20XX-XX-XX 20:40:00 & bedding/cushion/cushion\\
        \bottomrule
    \end{tabular}
}
\end{center}
\end{table}

In our work, sequence data account for a large portion of all our data, e.g., Rakuten Ichiba user shopping behavior data and Rakuten Points usage data. The shopping behavior data includes user purchases, canceled purchases and payment amounts data. All these behavior data are genre-based shopping records as introduced in the example table. The Rakuten Points data include user standard owned points, limited-term points and used points datasets. Although the data come from various domains and contain different internal information, we use this encoding strategy to embed the time-series related sequence data and standardize into image-like channel format. On the other hand, most users' data are very sparse in reality. The time interval between two shopping records can be very long. For the data storage and processing efficiency considerations, we use the vectorized data storage method to reduce the storage of redundant null values. For example, "2, 10, 28: 5000; 100, 20, 25: 6000; ..." it means that the user purchased the channel 28 item worth 5,000 at 10:00 on the 2nd day (relative time to one-year), then he purchased the channel 25 item worth 6000 at 20:00 on the 100th day.

\subsection{Meta-Profile Network}

\begin{figure*}[ht]
    \centering
    \includegraphics[width=\linewidth]{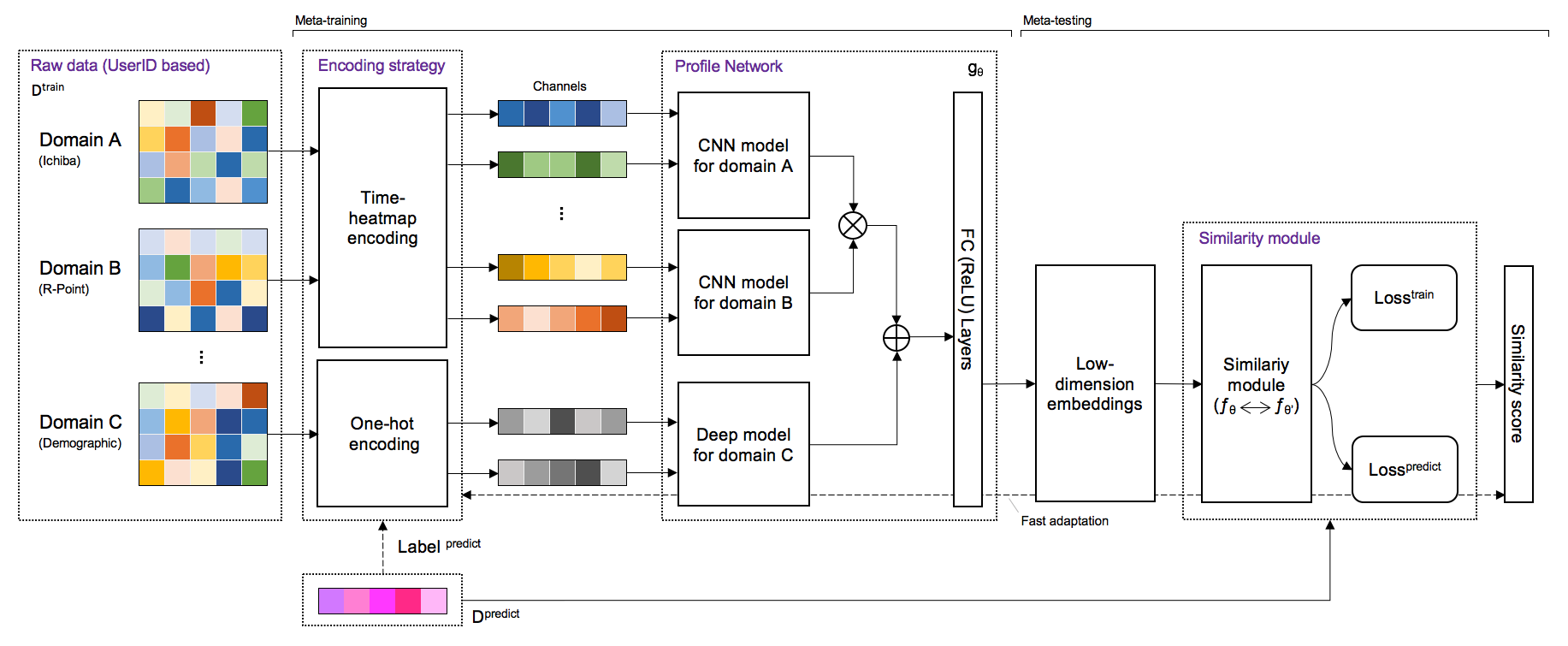}
    \caption{Meta-Profile Network architecture for N-way K-shot (N=9, K=1) problem with one prediction example case.}
    \label{fig:meta-profile-network}
\end{figure*}

To reduce the prediction error on unseen tasks, many studies point out that the model training process can be induced with a small support set, making the training process mimics what happens during inference, thus avoiding exposing all target labels to the model and encouraging the model to optimize quickly \citep{vinyals2016matching}. In our work, we define our source tasks as \(\tau = \{T_1, T_2, ..., T_n\}\), and unknown target task as \(\tau'\). We prepared three sets of data, i.e. a training set \(D^{train}\), a small support set \(S\), and a prediction set \(D^{predict}\). The training set is associated with the label space of \(\tau\) independently, while the support set and the prediction set share the same label space of \(\tau'\). We recognize each pair of support set \(S\) and mini-batch \(B\) sampled from training set as a data point \(d=\langle S,B \rangle\). For the case where the support set \(S\) contains \(K\) labeled examples for each of \(N\) classes, we consider it as a \(N\)-way \(K\)-shot classification task.

The training procedure has to be chosen carefully to match inference at test time \citep{vinyals2016matching}. We construct the Profile Network \(g_\theta\) (details of the deep network architecture described in Section 3.3) to learn to update model weights and optimize neural networks in each training iteration with embedded data point \(d\), and then encode the learned representations to feature maps \(g_\theta (D^{train},S)\). As metric-based meta-learning aims to learn a similarity measuring function over objects, the aim of our work is to learn such classifier for any given support set \(S\) in \(N\)-way \(K\)-shot classification. Similar to other metric-based models, the predicted probability over known tasks \(\tau\) is a weighted sum of labels of support set samples. The weight is generated by a similarity measuring function \(f_\theta\) with two data points. Consequently, the generated feature maps of \(g_\theta (D^{train},S)\) and \(g_\theta (D^{predict},S)\) are fed into similarity module \(f_\theta\) to produce a similarity score between each other. Theoretically, the similarity module \(f_\theta\) can be any kind of distance measurement kernel function. Here, we use a fully connected neural network with ReLU activation.

\subsection{Network Architecture}
This section introduces the deep network architecture of the profile network part in Figure.\ref{fig:meta-profile-network}.

\textbf{\textit{3.3.1 Justification of Model Selection.}}
The information derived from the user can be divided into two parts: intrinsic and extrinsic data. For instance, we define the sequential behavior data from services A and B as extrinsic data, and the demographic data from service C as intrinsic data. Regarding the extrinsic data (Figure.\ref{fig:meta-profile-network}), it is natural to employ a recurrent neural network (RNN) rather than a convolutional neural network (CNN) to tackle the sequential property. However, the advantage of RNN on sequential data relies on the assumption: items inside the sequence (e.g., words in NLP) have strong short-range and long-range dependence \citep{hochreiter1997long}. In practice, the dependence among different purchases from the same user is not strong. Two purchases of a user behavior sequence are approximately independent events. Besides, the frequency of user shopping behavior varies greatly. The huge difference in the sequence length for each user would cause problems in the implementation of the RNN model. Based on the above considerations, we use a CNN to model behavioral data converted into the image-like format, viz. data from domain A (Rakuten Ichiba) and B (Rakuten Points). 

\textbf{\textit{3.3.2 CNN Architecture for One Modality Model. }} 
Each CNN model has a 5-level structure, and each level of a convolutional layer consists of three parts in order: convolution, batch normalization and ReLU activation. Analogous to the division of the human life cycle, we expect each level of convolutional layer can periodically capture the spatial and temporal dependencies by following the patterns from 365 days, 52 weeks, 12 months, 4 seasons to 1 year. As each channel of input is a fixed length 8760 (365×24 = 8760 hours), we set the window size and max pooling size as Table.\ref{tab:cnn_model_setting} shows. 

\begin{table}[ht]
\caption{Table shows the details of the CNN architecture. In general, each convolutional layer consists of three parts in order: convolution, batch normalization and ReLU activation.}
\begin{center}
\label{tab:cnn_model_setting}
\resizebox{\linewidth}{!}{%
    \begin{tabular}{lll}
        \toprule
        \textbf{Layer Name} & \textbf{Output Shape} & \textbf{Model Settings}\\
        \midrule
        Concat\_0 & (32, 365, 24, ?) & Concatenate n-dim inputs\\
        \hdashline[0.5pt/2pt]
        Conv\_1 & (32, 365, 24, 64) & filters=64, kernel\_size=(7, 2), stride=1\\
        Max\_pool\_1 & (32, 52, 12, 64) & pool\_size=(7, 2)\\
        \hdashline[0.5pt/2pt]
        Conv\_2 & (32, 52, 12, 128) & filters=128, kernel\_size=(4, 2), stride=1\\
        Max\_pool\_2 & (32, 13, 6, 128) & pool\_size=(4, 2)\\
        \hdashline[0.5pt/2pt]
        Conv\_3 & (32, 13, 6, 256) & filters=256, kernel\_size=(3, 2), stride=1\\
        Max\_pool\_3 & (32, 4, 3, 256) & pool\_size=(3, 2)\\
        \hdashline[0.5pt/2pt]
        Conv\_4 & (32, 4, 3, 512) & filters=512, kernel\_size=(1, 1), stride=1\\
        Max\_pool\_4 & (32, 1, 1, 512) & pool\_size=(4, 3)\\
        \hdashline[0.5pt/2pt]
        Flatten & (32, 512) & Flatten\\
        \bottomrule
    \end{tabular}
}
\end{center}
\end{table}

\textbf{\textit{3.3.3 Source Task Settings. }} 
To learn the high-level abstraction of users in meta-training phase, we prepare two categories of tasks to induce the model to learn the universal meta information, viz., surveyed user profile information related tasks and various service usage related tasks. We use multi-task learning to compensate for the lack of labels in certain tasks. Based on the prior tests, we prune the number of source tasks from over one hundred to nine. 

Regarding the weighted losses of tasks in multi-task learning, we assign a weights to users depending on whether users are labeled for each task. We assign a weight 1 for true cases, otherwise 0. The overall loss for each mini-batch is shown below:
\begin{equation}
J(\theta)=\sum_{n=1}^{N} \sum_{i=1}^{B} \frac{w_{ni}}{\sum_{i=1}^{B}w_{ni}} J_{ni}(\theta)
\label{eq:weight_task}
\end{equation}
In Eq.\ref{eq:weight_task}, \(n\) denotes \(n\)-th task, \(i\) is the \(i\)-th user, \(w_{ni}\) is the weight, and it is 1 if user \(i\) has the label in the task \(n\) , and 0 otherwise. \(N\) is 9 in our case, \(B\) is the mini-batch size, which is chosen to be 32. \(J_{ni}(\theta)\) is the loss for user \(i\) in task \(n\) , and \(\theta\) is the model's parameters.

\section{Evaluation and Experiments}
In this section, we conduct experiments to evaluate the performance of our model including uncertainty and robustness estimates. The model has been implemented to the service called "Rakuten AIris" to provide effective target prospecting. We first explain the experiment settings. Then, we evaluate the effectiveness of our model with three experiments by comparing it with the baseline model. Finally, we briefly discuss the effectiveness of the model in online services. We will not reveal the details of data and tasks due to the data governance concerns.

\begin{table*}[ht]
\begin{center}
\caption{Out-of-distribution (OOD) detection performance with the normal training model (baseline model trained from scratch) and with the Meta-Profile Network model. AUROC is the Area Under the Receiver Operating Characteristic curve. AUPR is the Area Under the Precision-Recall curve. Results are based on 9-way 1-shot classification on 8 unseen target tasks.}
\label{tab:model_uncertainty}
\resizebox{\linewidth}{!}{%
    \begin{tabular}{l c c c c c c}
        \toprule
        & \multicolumn{3}{c}{\textbf{AUPR}} & \multicolumn{3}{c}{\textbf{AUROC}}\\
        \cmidrule(lr){2-4}\cmidrule(lr){5-7}
        \textbf{Target Task} & \textbf{Normal Training} & \textbf{Meta-Profile (diff)} & \textbf{Growth Rate} & \textbf{Normal Training} & \textbf{Meta-Profile (diff)} & \textbf{Growth Rate}\\
        \midrule
        RAKU\_BMW & 0.6166 & 0.6225(+0.0059) & 0.97\% & 0.8479 & 0.8597(+0.0118) & 1.39\%\\
        RAKU\_PRI & 0.4503 & 0.5361(+0.0858) & 19.06\% & 0.7006 & 0.7712(+0.0706) & 10.08\%\\
        RAKU\_MCD & 0.4470 & 0.4655(+0.0185) & 4.14\% & 0.7486 & 0.7587(+0.0101) & 1.35\%\\
        RAKU\_TYTP & 0.4456 & 0.4623(+0.0167) & 3.75\% & 0.7006 & 0.7250(+0.0243) & 3.47\%\\
        RAKU\_TYTA & 0.4200 & 0.4498(+0.0298) & 7.09\% & 0.6849 & 0.7220(+0.0371) & 5.42\%\\
        RAKU\_BK & 0.5498 & 0.5882(+0.0384) & 6.98\% & 0.7942 & 0.7968(+0.0026) & 0.33\%\\
        RAKU\_RB & 0.7041 & 0.7863(+0.0821) & 11.67\% & 0.8741 & 0.8914(+0.0173) & 1.98\%\\
        RAKU\_MB & 0.6891 & 0.7495(+0.0604) & 8.76\% & 0.8669 & 0.8927(+0.0259) & 2.98\%\\
        \bottomrule
    \end{tabular}
}
\end{center}
\end{table*}

\subsection{Experiment Setup}
\textbf{Datasets and tasks:} All of the experiments are performed on one-year of Rakuten e-commerce online user behavior data (Dec 1st, 2018 - Dec 1st, 2019). As explained in Section 3.1, our data come from three data sources: 1) user online shopping behavior sequence data from Rakuten Ichiba including item purchase and cancellation data, 2) point usage sequence data from Rakuten Point service, 3) inherent data such as demographic information. The goal of this work is to provide an efficient advertisement solution with representation learning for e-commerce platform users, the source tasks \(\tau\) are described in Section 3.3.3. The target tasks \(\tau'\) in meta-testing are business-related, based on the understanding of potential preference from user profiling, through a similarity check between existing users and newly developed users to provide an accurate advertisement recommendation. Here, 8 new unseen target tasks are tested. For the protection of personal privacy information, data anonymization and user pseudonymization processes are applied to all raw data. 

\noindent\textbf{Model settings:} The model is built and trained with Keras on top of TensorFlow using single GPU 'GeForce RTX 2080 Ti'. On average, initial meta-training takes about 15 hours with a single-core GPU. As for the CNN model settings, we choose the Adam optimizer, and apply the cyclic learning rate adaption strategy introduced in paper \citep{smith2017cyclical} to speed-up the convergence, and then use 'the Stochastic Gradient Descent with Warm Restarts' \citep{loshchilov2016sgdr} to further reduce the loss. Our multi-modal profile models show no overfitting during the training, and we think it is due to three reasons: 1) the strong regularization effect of the 2-D CNN; 2) multi-task learning; 3) a large training set. The number of model parameters is about 2 million. Based on the results of our experiments, we found our model performs well without using any additional regularization techniques like dropout or L1/L2 regularization.

\noindent\textbf{Baseline model}: To fairly evaluate the performance of Meta-Profile Network, all the following experiments are conducted by comparing the Meta-Profile Network to a normal training strategy. The normal training strategy completely copies the meta-testing part, which uses the exact same similarity module \(f_\theta\) and model parameter settings by training the prediction dataset \(D^{predict}\) from scratch.

\subsection{Uncertainty of Out-of-Distribution Detection}
The ability to distinguish abnormal data or data that is clearly far from current distribution space is a crucial indicator to judge whether the industrial model is successful, especially for deep learning and meta-learning models \citep{hendrycks2016baseline, hendrycks2019using}. To evaluate the performance of model in the uncertainty estimates, we consider using an out-of-distribution (OOD) detection test to see whether any out-of-distribution inputs are fed into the in-distribution classes. To measure the quality of out-of-distribution detection, two standard metrics Area Under the Receiver Operating Characteristic (AUROC) curve and Area Under the Precision-Recall (AUPR) curve are employed in our work. The higher AUPR and AUROC values indicate the stronger model's ability to resist uncertainty.

Table.\ref{tab:model_uncertainty} demonstrates the OOD detection performance of the baseline model and the Meta-Profile model. Since the distribution space of each task is different, we cannot just use a single unknown task to evaluate the uncertainty of the model. Here we test 8 different unseen target tasks in this estimation. Both the AUROC and AUPR scores of Meta-Profile Network consistently improve over the baseline model on all 8 unseen target tasks. For the most outstanding task 'RAKU\_PRI', Meta-Profile Network has more than 19\% AUPR and 10\% AUROC performance improvement in the uncertainty test. Compared to the normal training strategy, Meta-Profile Network has around 8\% AUPR and more than 3\% AUROC improvement in 8 tasks on average. Consequently, Meta-Profile Network can significantly improve the performance of the model in resisting uncertainty compared to the normal training strategy.

\subsection{Robustness to Data Insufficiency (Label Masking)}
In practice, the data insufficiency problem is particularly prevalent. Like the fuel to the engine, the success of a deep model is based on having enough sufficient labeled samples. In extreme cases, when we have only a few samples for a task, the effectiveness of the model will be greatly reduced. Meta-learning has demonstrated that it can mitigate the side effects of the data insufficiency problem. We design a label masking experiment to evaluate the robustness of our model when it faces the insufficient data problem. 

\begin{figure}[ht]
    \centering
    \includegraphics[width=\linewidth]{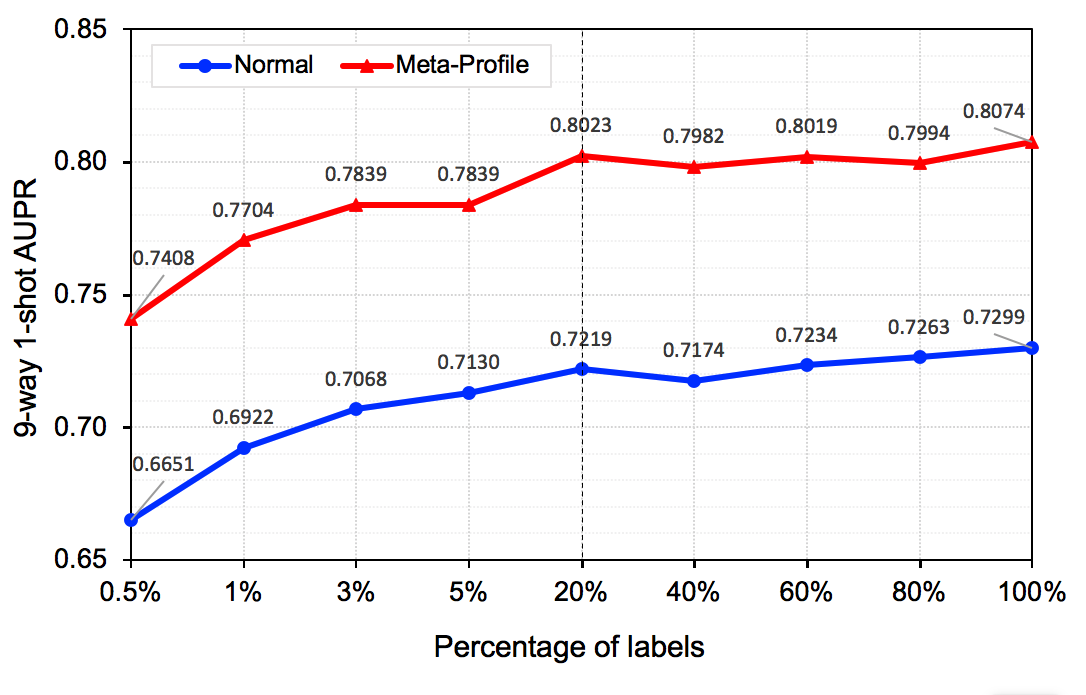}
    \caption{Robustness evaluation in a data insufficient scenario. To simulate a situation with data insufficiency, the original labeled samples are manually masked by percentage. We compare the performance of the two models under the moderate and extremely low-resource situations. The value of x-axis represents the percentage of labels available.}
    \label{fig:robustness_label_masking}
\end{figure}

To quantitatively observe and analyze the performance of the model in different levels of masking, we prepare two sets of test cases to simulate two insufficient data scenarios, a moderate scenario and an extreme scenario. The original size of unknown labeled samples is 10,000. The data size of the moderate group gradually decreases from 100\% in 20\% decrements to 20\%. The extreme group tests the performance of the model under severely reduced datasets with, 5\%, 3\%, 1\%, 0.5\%, of labels respectively. Figure.\ref{fig:robustness_label_masking} shows the performance of two models in a data insufficiency scenario. Compared with the normal training strategy, Meta-Profile Network shows relatively good performance in all test cases. Even in the case of extremely small data, our model can still improve task prediction performance by 10\%. Similar results are demonstrated in other pre-trained model related studies under low-resource conditions \citep{bhunia2019handwriting}.

\subsection{Robustness to Class Imbalance}
In terms of model robustness evaluation, another common problem that needs attention is the imbalance of class labels. Most machine learning algorithms perform best with an equal number of class samples. In cases where the total number of samples from a class is more abundant than samples from other classes, it will induce the model to learn more about the prevalent class and produce a result that looks good but is far from the ground truth. Understanding the differences in the learning of new tasks caused by the unequal distribution of minority and majority classes is important to this work.

\begin{figure}[ht]
    \centering
    \includegraphics[width=\linewidth]{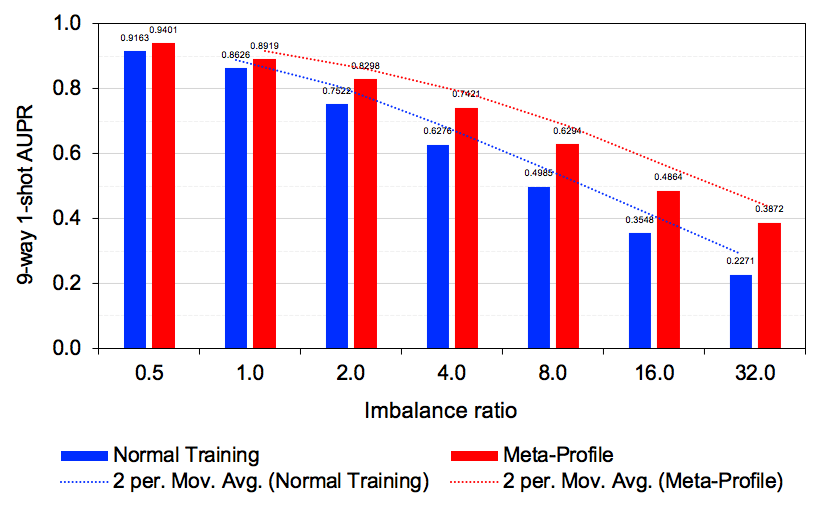}
    \caption{Robustness evaluation in a class imbalance scenario. The imbalance ratio is defined as the ratio of positive class to negative class. We prepare 7 test cases with an exponentially increasing imbalance ratio starting at 0.5. Dashed lines indicate the 2-period moving average. }
    \label{fig:robustness_class_imbalance}
\end{figure}

An additional explanation to the task description in Section 4.1, for the new brand advertisement recommendation task, we set the newly developed users as the positive class. In contrast, users that never used this brand before are considered the negative class (these users are not involved in the training set \(D^{train}\)). The imbalance ratio is defined as the ratio of positive class to negative class. By adjusting the imbalance ratio, we can test the performance of the model under class imbalanced circumstances. We test 7 cases with an exponentially increasing imbalance ratio starting at 0.5. Compared to the baseline model, the figure illustrates (Figure.\ref{fig:robustness_class_imbalance}) that Meta-Profile Network significantly improves the performance with the imbalance ratio increasing. Meanwhile, our model performs better compared to normal training as the imbalance ratio increases, as shown by the slope of the line representing the 2-period moving average of AUPR. According to the results of two robustness estimates, our model can significantly improve the robustness score compared to the normal training strategy, which can help the application cope with more extreme conditions.

\subsection{Model Deployment}
\begin{figure}[ht]
    \centering
    \includegraphics[width=0.5\linewidth]{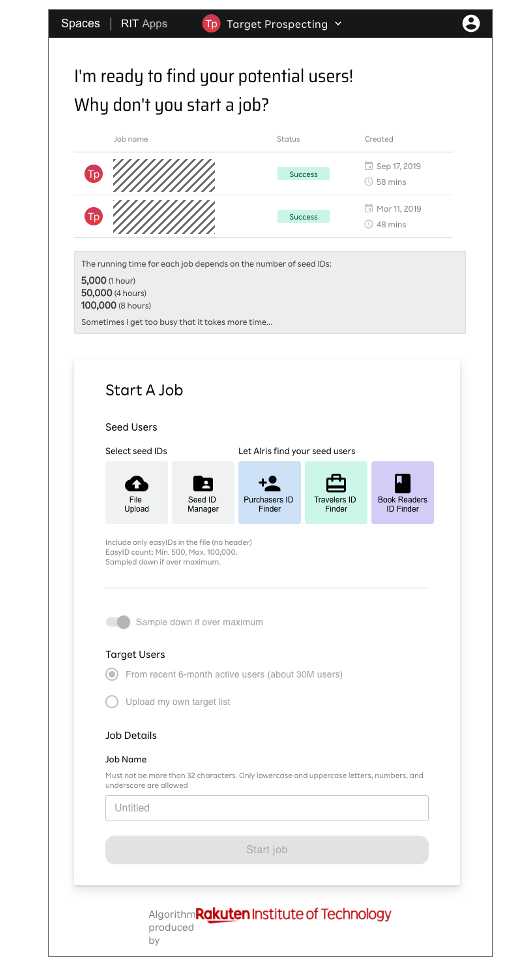}
    \caption{Rakuten AIris target prospecting service page. The proposed User Meta-Profile Network running in its background enables fast adaptation to new use cases, instead of training models from scratch.}
    \label{fig:airis_web}
\end{figure}

Lastly, we have deployed our Meta-Profile Network online on Rakuten AIris advertisement recommendation service (Figure.\ref{fig:airis_web}). The service allows users to do the target prospecting without training the entire model from scratch when we have a new task (job), e.g., find the potential users with similar behavior patterns to new brand's existing users. According to the statistics, the customer acquisition cost (CAC) for AIris target prospecting has a 30-50\% decrease compared to the standard data management platform (DMP) segments or other platform's lookalike solution. The customer conversion rate (CVR) has a high improvement from 10-30\% in comparison with the previous standard solution, which is considered to be a big improvement for the service.

\section{Conclusion}
In this work, we introduced the Meta-Profile Network, a novel meta-learning framework for learning how to profile user behaviors and fast adapt to subsequent learning tasks. The proposed time-heatmap encoding strategy, multi-modal combined with multi-task learning architecture proved to be highly efficient for learning representation of user behavior with long-term cross-domain data. Meta-Profile Network based on metric-based meta-learning and deep neural networks empower the model to effectively and quickly transfer the learned knowledge to unknown tasks while learning a good profile of user behaviors.  

Our extensive experiments demonstrate that the benefits of the Meta-Profile Network extend beyond merely model accuracy and quick convergence in comparison with training from scratch strategy. In fact, the proposed meta-learning approach has better robustness and uncertainty under various extreme conditions as well. Moreover, Meta-Profile Network also shows strong flexibility and scalability in practice, which could promote meta-learning research and development in large-scale industrial applications.

\begin{acks}
This work is supported by Rakuten Institute of Technology, Tokyo, Japan. We are grateful to cooperate with the Rakuten AIris science team. The authors would also like to thank the anonymous reviewers for their valuable comments and helpful suggestions.
\end{acks}

\balance
\bibliographystyle{unsrt}
\bibliography{main}
\end{document}